\pdfoutput=1
\documentclass[letterpaper]{article} %
\usepackage{aaai23}
\usepackage{times}  %
\usepackage{helvet}  %
\usepackage{courier}  %
\usepackage[hyphens]{url}  %
\usepackage{graphicx} %
\usepackage{natbib}  %
\usepackage{caption} %
\usepackage{algorithm}
\usepackage{algorithmic}
\usepackage{amsmath}
\usepackage[table,xcdraw]{xcolor}
\usepackage{float}

\usepackage{makecell}
\usepackage{xcolor}
\usepackage{xspace}

\usepackage{amsmath}
\usepackage{cleveref}
\usepackage{amssymb}
\usepackage{diagbox}

\usepackage{newfloat}
\usepackage{listings}
\DeclareCaptionStyle{ruled}{labelfont=normalfont,labelsep=colon,strut=off} %
\floatstyle{ruled}
\newfloat{listing}{tb}{lst}{}
\floatname{listing}{Listing}
\title{Building Coverage Estimation with Low-resolution Remote Sensing Imagery}
\author {
    Enci Liu, Chenlin Meng, Matthew Kolodner, Eun Jee Sung, Sihang Chen\\Marshall Burke, David Lobell, Stefano Ermon
}
\affiliations {
    \vspace{1mm}
    Stanford University\\
    \vspace{2mm}
    \{chenlin, jesslec\}@cs.stanford.edu,
    \{mkolod, ejsung, schen22, mburke, dlobell\}@stanford.edu,
    ermon@cs.stanford.edu
}

\usepackage{bibentry}

\begin{document}

\maketitle

\begin{abstract}
    Building coverage statistics provide crucial insights into the urbanization, infrastructure, and poverty level of a region, facilitating efforts towards alleviating poverty, building sustainable cities, and allocating infrastructure investments and public service provision. Global mapping of buildings has been made more efficient with the incorporation of deep learning models into the pipeline. However, these models typically rely on high-resolution satellite imagery which are expensive to collect and infrequently updated. As a result, building coverage data are not updated timely especially in developing regions where the built environment is changing quickly. In this paper, we propose a method for estimating building coverage using only publicly available low-resolution satellite imagery that is more frequently updated. We show that having a multi-node quantile regression layer greatly improves the model's spatial and temporal generalization. Our model achieves a coefficient of determination ($R^2$) as high as 0.968 on predicting building coverage in regions of different levels of development around the world. 
    We demonstrate that the proposed model accurately predicts the building coverage from raw input images and generalizes well to unseen countries and continents, suggesting the possibility of estimating global building coverage using only low-resolution remote sensing data.
\end{abstract}

\section{Introduction}
The quantity and location of buildings provide important insight into the human activities and urban development of a region. Not only are building statistics themselves key socioeconomic indicators, they also help predict other key sustainable development indices, including poverty 
\cite{Ayush2021EfficientPM,uzkent2020efficient,Yeh2020UsingPA},
population density \cite{huang2021100}, and climate outcomes \cite{chini2018towards}. Moreover, building coverage statistics help policymakers and NGOs make informed decisions regarding the provision of public services, the targeting of humanitarian aid, and priorities for large-scale infrastructure investments.

The development of deep learning detection \cite{redmon2018yolov3} and segmentation models \cite{ronneberger2015u,sun2019deep} has allowed for more efficient global mapping of buildings. As a result, there has been an increasing number of global settlement map datasets in the past decade \cite{esch2017breaking,marconcini2020outlining,sirko2021continental}, allowing researchers to develop insights into the socioeconomic development of different regions.

Nevertheless, detection- or segmentation-based methods typically rely on a large amount of high-resolution satellite imagery and the corresponding pixel- or instance-level labels for training, which are often prohibitively expensive and unaffordable for researchers and policymakers \cite{meng2022count}. Moreover, high-resolution imagery are updated less frequently than low-resolution ones. In addition, running detection or segmentation models over high-resolution images covering a large area requires a large amount of compute. Due to these reasons, building data gathered in this way is often not updated timely. For example, the Microsoft Global Building Footprints dataset\footnote{\url{https://github.com/microsoft/GlobalMLBuildingFootprints}} is collected from satellite images between 2014 and 2021. However, during the 7-year period, population continues to grow and new buildings are constructed, especially in fast developing regions. For instance, from 2015 to 2020, the population increased by 44.1\% for Malappuram and 34.2\% for Abuja.\footnote{\url{https://worldpopulationreview.com/}} Moreover, fine-grained detection and segmentation models usually generalize poorly to unseen geographies, timestamps, and image sources because the appearance of buildings vary widely in satellite images \cite{yuan2014learning}.

Compared with its high-resolution counterpart, low-resolution satellite imagery (e.g. Sentinel-1 and Sentinel-2) are publicly available and updated every month, making them desirable for studying the economic and urban development of a region. However, prior works have not fully utilized low-resolution imagery. 
In this paper, we propose a cheaper and more generalizable way to update building statistics using only low-resolution satellite imagery from Sentinel-1 and Sentinel-2.
Instead of detecting or segmenting each building in satellite images, the proposed model directly predicts the building coverage from the raw input pixels, using input imagery from a public source that is updated nearly weekly. 
Specifically, we found that incorporating a multi-node quantile regression loss helps improve the generalization of the model.
The proposed method achieves a coefficient of determination ($R^2$) as high as $0.968$ in regions from different continents and of different levels of development. We also conduct ablation studies and show that the incorporation of additional multi-spectral bands available from the low-resolution satellite imagery and the multi-node quantile regression design help improve the model performance. 

\section{Method}
In this paper, we propose a deep-learning based regression model that accurately predicts building coverage in low-resolution satellite imagery from Sentinel-1 and Sentinel-2. Unlike detection- or segmentation-based methods, our model does not require high-resolution training data or hand-crafted threshold and generalizes well to unseen regions. The proposed method accurately predicts the building coverage in the raw input low-resolution satellite image with the help of a quantile regression.

\begin{figure*}
\begin{center}
\includegraphics[width=1.0\linewidth]{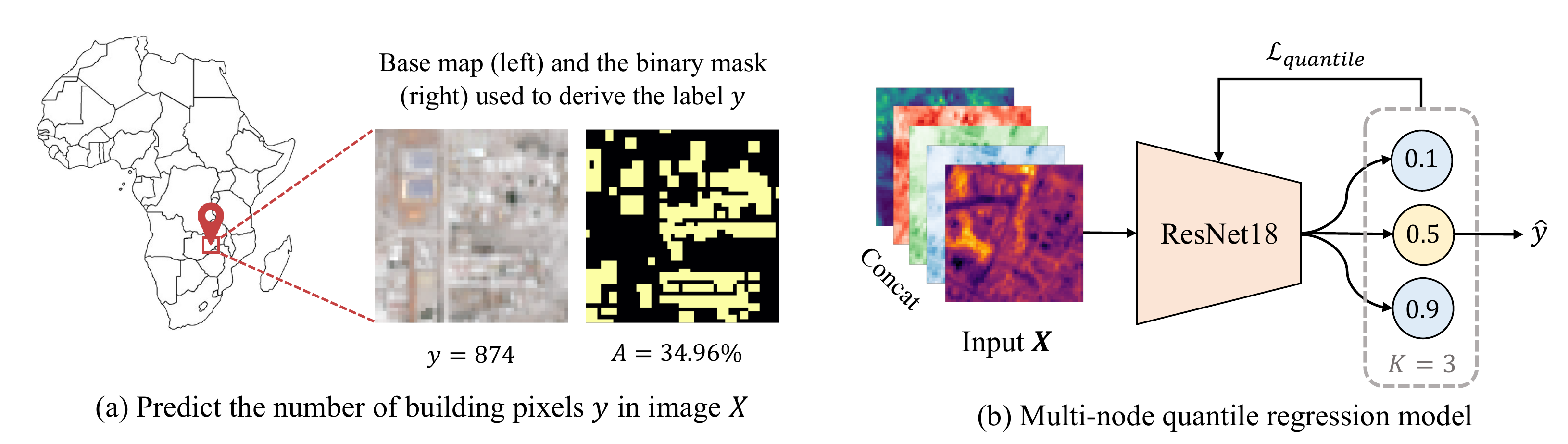}
\end{center}
   \caption{(a) Given an input image $X$, we want to predict the number of building pixels $y$ within it. We assume that $y$  approximate the actual building coverage. (b) Architecture of the multi-node quantile regression model. The input to the model includes five channels, which are Sentinel-1 band 1 (Overall Mean), Sentinel-2 band 4, 3, 2, 8 (R, G, B, NIR). The model has $K$ output nodes representing different quantiles.
   }
\label{fig:model}
\end{figure*}

\subsection{Problem Definition}
Given a geographical region, we want to estimate the building coverage within it. Due to the lack of direct statistics of building coverage in square meter or kilometer, we instead predict the number of building pixels $y \in \mathbb{R}$, in the satellite image $X$ representing the target region, assuming that the number of building pixels is proportional to the actual building coverage (Figure \ref{fig:model}(a)). 
We want to build a model that predicts $y$ from the raw image input $X$. 

\subsection{Multi-node Quantile Regression Model}
We observe that the distribution of building pixel counts across Africa and South America are heavy-tailed, with more than 75\% of the samples having less than 20\% of all pixels being buildings. Regular linear regression trained on root-mean-square error objective estimates the conditional mean and assumes normality of the data distribution, failing to model non-normal asymmetric distribution accurately. Moreover, linear regression is not robust to outlier values, which characterize the building pixel count distribution for our task. 

To address the aforementioned problems, we adopt quantile regression \cite{koenker1978regression}, which estimates the conditional quantile (e.g., median) of the response variable. Quantile regression allows us to incorporate uncertainty in prediction and captures the relationship between the input and different quantiles of the data. As we will see in Section \ref{results}, multi-node quantile regression indeed empirically performed better than regular linear regression for our task.

Specifically, we modify the ResNet18 \cite{he2016deep} architecture to have $K$ output channels (i.e. nodes), each corresponding to a different quantile (see Figure \ref{fig:model}). 
As it is observed that the median of a distribution gives the minimum absolute error from the ground truth \cite{hanley2001visualizing}, at inference time, we collect the model predictions from only the 0.5 quantile node, which is expected to predict the conditional median of the response variable.

\subsection{Multi-node Quantile Loss}
For each node that represents a quantile $q \in (0, 1)$, we compute an asymmetric quantile loss, or pinball loss. Depending on the quantile $q$, over- and under-estimation are penalized unevenly. 
Specifically, the node-wise pinball loss for a given prediction $\hat{y}$ and ground truth label $y$ is computed as follows:
\[ \mathcal{L}_\text{pinball}(q, y, \hat{y}) = 
\begin{cases} 
      q \cdot |\hat{y} - y| & \text{if } \hat{y} \ge y \\
      (1 - q) \cdot |\hat{y} - y| & \text{otherwise}
   \end{cases}
\]
When $q=0.5$, the pinball loss is the same as the absolute error. 

To compute the final loss, we take the mean of the pinball losses among the $K$ output nodes as follows:
\begin{equation*}
    \mathcal{L}_\text{quantile}(y, \hat{y}) = \frac{1}{K}\sum_{n=1}^{K} \mathcal{L}_\text{pinball}(q_n, y, \hat{y})
\end{equation*}
where the subscript $n$ indicates the index of the quantile in the list of quantiles predicted by the model. Notice that when $K=1$ and $q=0.5$, the quantile loss formula boils down to the regular $L_1$ loss. 
\section{Experiment Setup}

\begin{figure}[t]
\begin{center}
\includegraphics[width=1.0\linewidth]{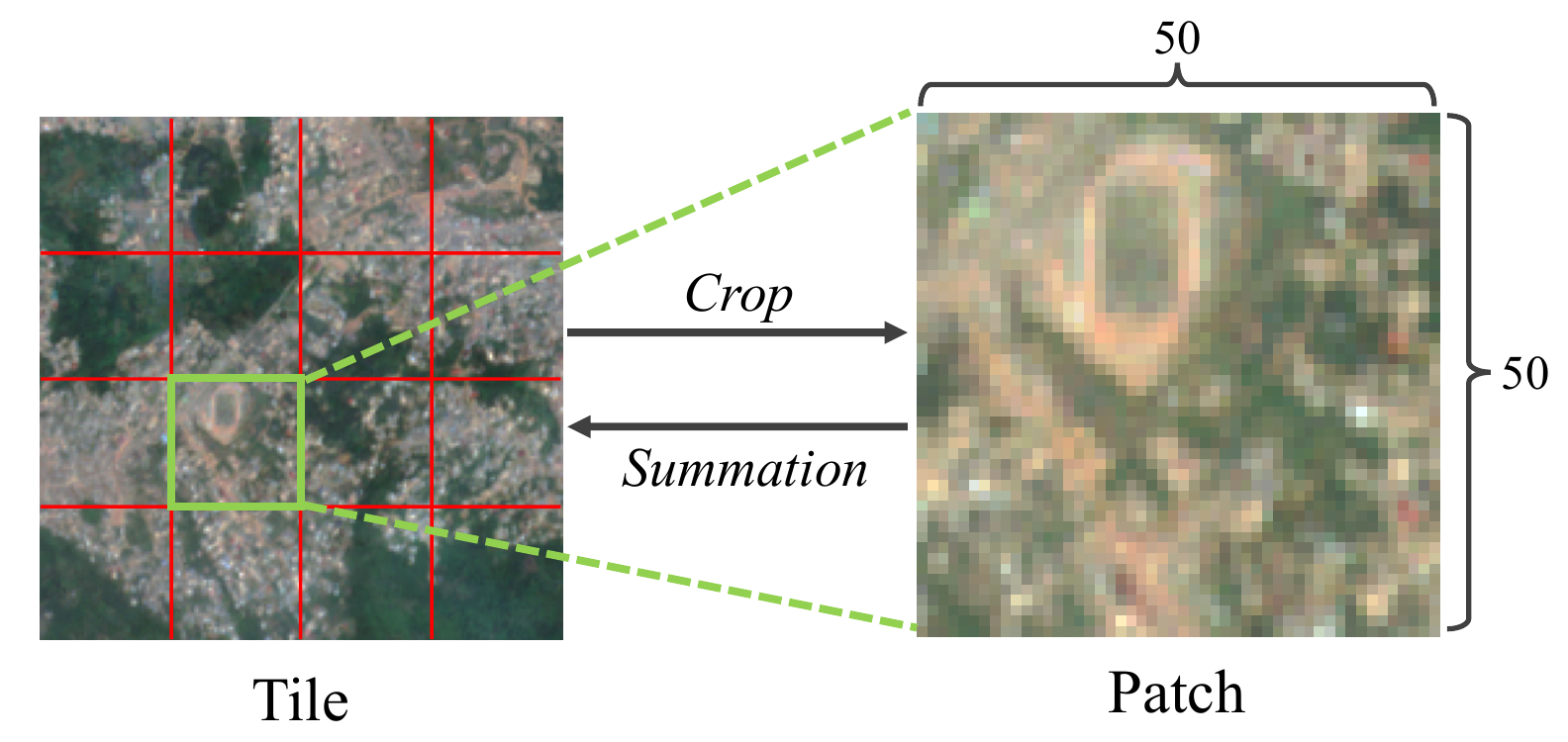}
\end{center}
   \caption{A \textbf{tile} can be cropped into multiple $50\times50$ pixels \textbf{patches}. The \textbf{tile-level} results are computed by taking summation of all the \textbf{patch-level} predictions.
   }
\label{fig:visualization}
\end{figure}

Before introducing the experiment setups, we define \textit{tile} and \textit{patch} in the context of this paper. We define a \textit{patch} to be the small rasters of $50\times 50$ pixels that we crop larger rasters into. 
A \textit{tile} is a larger satellite imagery that could be cropped into multiple smaller \textit{patches} (see Figure \ref{fig:visualization}). We train and evaluate the multi-node quantile regression model on patches of $50\times50$ pixels and add post-processing steps to collect tile-level results. 

\subsection{Training Data}
As the majority of the African continent is covered by forest or desert and does not contain any buildings, we sample locations based on the population density
so that the training data contains sufficient tiles with buildings for the model to learn from. Our training set contains 15,000 input-label pairs.
Sentinel-1 and Sentinel-2 images are used as as the input data and the Open Buildings dataset is used to derive the building coverage label, as we detail next.

\paragraph{Sentinel-1 (S1)} satellites collect radar imagery for land and ocean monitoring.
We download the S1 satellite imagery collected in 2020 from Google Earth Engine.
The images have 10m GSD and are composites that take the median value over the target period of time. 
We include band 1, the VV overall mean, as one of the input channels. The S1 tiles are cropped into $50 \times 50$-pixel patches. As areas with high built-up density are shown to result in stronger signals in S1 band 1 \cite{koppel2017sensitivity}, we expect that adding this channel to the input will improve the model's performance.

\paragraph{Sentinel-2 (S2)} satellites are equipped with the mission of land monitoring.
We download the 2020 composites of S2 imagery from Google Earth Engine.
We include the RGB channels (i.e. bands 4, 3, 2) and the near-infrared (NIR) channel (i.e. band 8) from S2 as input. Compared with other channels, NIR is useful for distinguishing the vegetation from the buildings~\cite{luo2019fusing, pessoa2019photogrammetric,  schlosser2020building}. All of the four bands are available in 10m GSD. The S2 tiles are cropped into 50 $\times$ 50-pixel patches.

\paragraph{Open Buildings} \cite{sirko2021continental} contains 516M building footprints across 43 African countries that cover 64\% of the continent. The building footprints were detected using state-of-the-art segmentation model collected from \emph{high-resolution imagery} at different timestamps. We downsample the original high-resolution mask (0.5m GSD) to 10m GSD to match the resolution of the input data. Then, we convert the continous-valued rasters into binary masks by thresholding at 0.
The building pixel count labels are derived for the $50 \times 50$-pixel patches.

\subsection{Experiment Settings}
To evaluate the generalization performance of the model, we consider three experiment settings that captures likely cases of real-world application. 

\paragraph{Holistic.} In the holistic setting, we train and test on data points sampled across the African continent based on population density.

\paragraph{Intra-country.} In the intra-country setting, we train and test the model on samples from the same country. 

\paragraph{Exclusive.} In the exclusive setting, we train the model on all African countries except for the one country on which we test our model.

\subsection{Baselines}
To evaluate the performance of the proposed method, we use existing settlement map products from different years as baselines to compare against. These off-the-shelf products provide researchers and policymakers with general information about urban shapes and boundaries, but could be less useful in providing up-to-date building coverage statistics, which change more frequently than the shape of urban area.
We believe that these existing settlement map products serve as good baselines to compare our method against and provide information of them in this section.

\paragraph{Gloal Urban Footprint (GUF)} 
GUF was collected from 2011 to 2012. It contains mappings of human settlements in the form of binary masks. 

\paragraph{Global Human Settlement Layer (GHSL)}
GHSL was collected from Sentinel-1 images in 2018. The building map is available as binary masks.

\paragraph{World Settlement Footprint (WSF)}
WSF \cite{marconcini2020outlining} was collected in 2015. It contains binary masks of global human settlements.

\begin{figure*}[t]
\begin{center}
\includegraphics[width=1.0\linewidth]{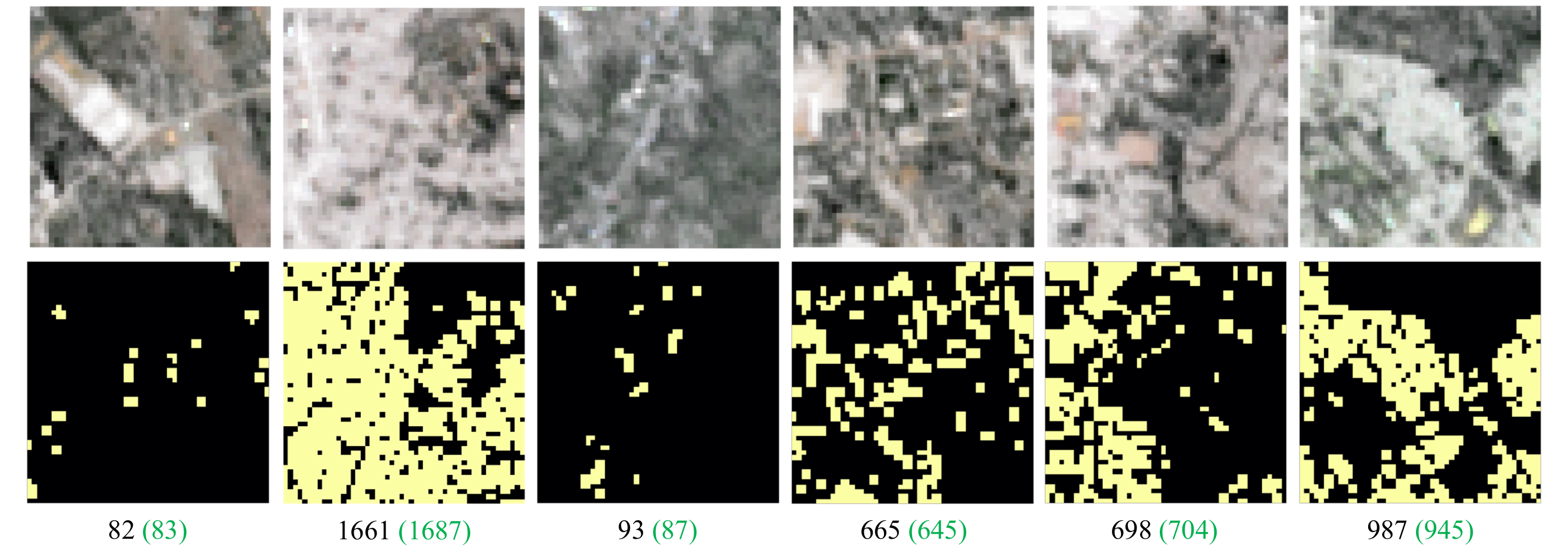}
\end{center}
   \caption{Examples of model predictions in Brazil, which was unseen during training. The first row is the RGB input image; the second row is the binary masks (where the bright yellow pixels are building pixels) from which we derive the label. The model prediction is in black; the ground truth labels are highlighted in green in the parentheses. Our method accurately estimated the results.
}
\label{fig:success_examples}
\end{figure*}

\subsection{Evaluation Settings}
We evaluate the model performance at both patch-level and tile-level and describe the evaluation metrics in this section.

\subsubsection{Patch-level Evaluation} 
As the model is trained on patches, we want to evaluate the model's performance at the same scale. We use two evaluation metrics for patch-level evaluation: mean absolute error (MAE) and Pearson's $r^2$ between the predicted and the ground truth labels.

\subsubsection{Tile-level Evaluation} 
In real-world applications, building coverage statistics are needed over a large geography. To reflect this use case, we also evaluate the model performance at tile-level using absolute error in building coverage. To compare the statistics of building coverage across baselines with different GSDs, we compute the percentage of building pixels within a tile as a proxy for building coverage. For a tile $R$ cropped into $N$ patches at inference time, let $y_i$ be the number of building pixels in patch $i$, the building coverage percentage is computed as:
\begin{equation*}
    C_{building}(R) = \frac{\sum_{i=1}^{N} y_i}{H_{tile} \times W_{tile}} \times 100
\end{equation*}
where $H_{tile}$ and $W_{tile}$ denote the height and width (in pixel) of the tile. The absolute error between a method and the ground truth is computed as the absolute difference between the two building coverage percentages.

\begin{table}[]
\centering
\begin{tabular}{lcc}
\Xhline{3\arrayrulewidth}
\textbf{Expt./Eval. settings}* & \textbf{Open Buildings} & \textbf{SpaceNet7} \\ \hline
Holistic            &  \checkmark    &           \\
Intra-country       &  \checkmark   &           \\
Exclusive           &  \checkmark    &           \\ \hline
Patch-level         &   \checkmark   & \checkmark \\
Tile-level          &                &  \checkmark\\ \Xhline{3\arrayrulewidth}
\end{tabular}
\caption{Checkmarks indicate that the corresponding dataset is used as validation data under the experiment (Expt.) or evaluation (Eval.) settings. *Different experiment settings require retraining the model, while different evaluation settings do not.
}
\label{tab:evaluation_data}
\end{table}

\subsection{Evaluation Data}
We evaluate our model on Open Buildings and SpaceNet 7 Challenge datasets and provide the information in Table \ref{tab:evaluation_data}.

\subsubsection{Open Buildings}
The Open Buildings \cite{sirko2021continental} dataset provides the training labels. We evaluate the model performance on a hold-out test subset of the Open Buildings dataset at \textit{patch-level} only. We do not use Open Buildings for tile-level evaluation because it contains data from different timestamps.

\subsubsection{SpaceNet 7 Challenge dataset (SpaceNet7)}
SpaceNet7\footnote{SpaceNet on Amazon Web Services (AWS). “Datasets.” The SpaceNet Catalog. Last modified October 1st, 2018. Accessed on November 20th, 2021. \url{https://spacenet.ai/datasets/}} was published in 2020 as the data for the SpaceNet 7 Multi-Temporal Urban Development Challenge. The dataset provides $4 \text{km} \times 4$km tiles and the building polygons in each tile. 
We downsample the raster to 10m GSD to match the resolution of the input data. We evaluate the model performance on SpaceNet7 at both \textit{patch-level} and \textit{tile-level}.

We use the SpaceNet7 tiles for validation because the labels were collected the same year as the Sentinel-1/-2 input data. Furthermore, SpaceNet7 includes tiles from regions outside of Africa, allowing us to evaluate the model's performance on unseen countries. Specifically, we evaluate our model on six SpaceNet7 tiles from different regions: Uganda, Zambia, Ghana, Peru, Brazil, and Mexico. Note that we do not use SpaceNet7 as the training labels because the data is limited in quantity. These labels are human-generated and likely of higher quality compared to those from Open Buildings, which are generated by a model.

\section{Results}\label{results}

\begin{table*}[h]
\small
\centering
\begin{tabular}{ccccccc|cccccc}
\Xhline{3\arrayrulewidth}
\multicolumn{1}{l}{}                              & \multicolumn{6}{c|}{\textbf{Africa}}                                                                                                                  & \multicolumn{6}{c}{\textbf{South America}}                                                                                                                  \\ \hline
\multicolumn{1}{l}{}                              & \multicolumn{2}{c}{Uganda}                                & \multicolumn{2}{c}{Zambia}                                & \multicolumn{2}{c|}{Ghana} & \multicolumn{2}{c}{Brazil}                                & \multicolumn{2}{c}{Peru}                                  & \multicolumn{2}{c}{Mexico} \\ \hline
Method                                          & $R^2\uparrow$ & Tile $\downarrow$                                              & $R^2\uparrow$ & Tile $\downarrow$                                             & $R^2 \uparrow$        & Tile $\downarrow$       & $R^2\uparrow$ & Tile $\downarrow$                                            & $R^2\uparrow$ & Tile $\downarrow$                                             & $R^2\uparrow$         & Tile $\downarrow$       \\ \hline
\multicolumn{1}{c|}{GUF (2012)}                   &  0.092     &  \multicolumn{1}{c|}{7.23}                             & -0.715      & \multicolumn{1}{c|}{\textbf{0.96}}                             &   0.466          &    1.59         & --    & \multicolumn{1}{c|}{--}                             &   --    & \multicolumn{1}{c|}{--}                             &     --          &   --          \\
\multicolumn{1}{c|}{WSF (2015)}                   & 0.286      &  \multicolumn{1}{c|}{\textbf{0.21}}                             &  -5.444     & \multicolumn{1}{c|}{38.68}                             &  -0.290           &   3.28          &  0.579      & \multicolumn{1}{c|}{9.21}                             & 0.562      & \multicolumn{1}{c|}{6.06}                             & -0.099              &  18.52           \\
\multicolumn{1}{c|}{GHSL (2018)}                  & 0.057     &  \multicolumn{1}{c|}{6.83}                             & 0.023     & \multicolumn{1}{c|}{2.46}                             &  0.771            &  0.75           &  0.863     & \multicolumn{1}{c|}{0.97}                             &  0.516     & \multicolumn{1}{c|}{10.06}                             &   0.187          & 8.12            \\\hline
\multicolumn{1}{c|}{w/o multi-node QR} & -15.51 & \multicolumn{1}{c|}{33.15} & -10.41 & \multicolumn{1}{c|}{54.27} & -15.64        & 31.68        & -0.069 & \multicolumn{1}{c|}{18.30} & -2.807 & \multicolumn{1}{c|}{38.28} &   -2.480       & 36.24        \\
\multicolumn{1}{c|}{w/o S1 band 1} & 0.809 & \multicolumn{1}{c|}{1.95} & 0.779 & \multicolumn{1}{c|}{3.74} & 0.252        & 3.37       & 0.864 & \multicolumn{1}{c|}{3.91} & \textbf{0.906} & \multicolumn{1}{c|}{\textbf{2.34}} & 0.422        & 10.96       \\
\multicolumn{1}{c|}{w/o S2 band 8} & 0.772 & \multicolumn{1}{c|}{2.33} & 0.580 & \multicolumn{1}{c|}{7.14} & 0.804        & \textbf{0.54}        & 0.751 & \multicolumn{1}{c|}{6.19} & 0.836 & \multicolumn{1}{c|}{3.13} & 0.337         & 13.36        \\
\multicolumn{1}{c|}{Ours} & \textbf{0.868} & \multicolumn{1}{c|}{1.18} & \textbf{0.866} & \multicolumn{1}{c|}{3.21} & \textbf{0.835}        & 2.31        & \textbf{0.968} & \multicolumn{1}{c|}{\textbf{0.83}} & 0.798 & \multicolumn{1}{c|}{6.63} & \textbf{0.707}         & \textbf{0.36}        \\ \Xhline{3\arrayrulewidth}
\end{tabular}
\caption{Results on SpaceNet7. The bottom four rows are the proposed methods with the corresponding components removed and the complete version. In the table, $\mathbf{r^2}$ is the patch-level Pearson's $r^2$ and \textbf{Tile} is the tile-level absolute error between SpaceNet7 and the corresponding method. The model was trained with 15,000 samples in Africa for 1200 epochs and tested on the corresponding SpaceNet7 tiles. 
}
\label{tab:spacenet_results}
\end{table*}

\begin{table}[t]
\centering
\begin{tabular}{c|c|c|cc}
\Xhline{3\arrayrulewidth}
\textbf{Expt. setting}    & \textbf{Train}   & \textbf{Test} & \textbf{MAE} $\downarrow$ & $\mathbf{R^2}\uparrow$ \\ \hline
Holistic            & Africa                  & Africa               & 75.43      & 0.888          \\
Intra-country       & Rwanda                  & Rwanda               & 27.60      & 0.938         \\ 
Exclusive  & Africa* & Rwanda               & 42.04        &  0.844     \\
Exclusive & Africa* & Uganda      & 207.74         & 0.568           \\
Excluisve & Africa*  & Kenya       & 110.67        & 0.915          \\ \Xhline{3\arrayrulewidth}
\end{tabular}
\caption{Patch-level results for different experiment settings on Open Buildings. All models in the table are trained with 15,000 samples from the corresponding train regions for 1200 epochs and tested on 1,000 samples from the corresponding test regions. *Under the Exclusive setting, the corresponding test region is removed from the training set so that the model is tested on unseen regions.}
\label{tab:open_buildings_results}
\end{table}

In this section, we evaluate the performance of the proposed method. We first show that our model accurately predicts building coverage in Africa and generalizes to South American regions unseen during training. We also conduct ablation studies demonstrating that the major design choices -- non-RGB bands and multi-node quantile regression -- are necessary for boosting model performance and generalization.

\subsection{Accurate Prediction in Africa}
To demonstrate the effectiveness of our method, we evaluate it on both SpaceNet7 (see Table \ref{tab:spacenet_results}) and Open Buildings (see Table \ref{tab:open_buildings_results}) in Africa. We define a \textit{tile} as multiple \textit{patches} and compute the tile-level results by taking summation of the patches within it (see Section 4.1).

As shown in Table \ref{tab:spacenet_results}, our model achieves an $R^2$ as high as $0.968$ at patch-level, outperforming the baselines on most of the African regions. Some examples of the proposed model's patch-level prediction are provided in Figure \ref{fig:success_examples}. Furthermore, the proposed method yields fairly accurate building coverage estimates at tile-level, achieving a low error rate of 0.54\% on Ghana. Additionally, we observe that baselines like WSF and GUF give good estimates at only tile-level but not the other, indicating that the errors at patch-level are cancelled out. In contrast, the proposed method yields consistently accurate estimations at both tile- and patch-level.
We emphasize that having good performance on both scales is ideal, since it gets us closer to finer-grained pixel-level predictions (semantic segmentation).

\subsection{Generalization to Unseen Regions}
Prior methods for generating building coverage statistics generalize poorly under domain shift. To evaluate the generalizability of the proposed method, we train and test the multi-node quantile regression model under three experiment settings (defined in Section 4.3) -- Holistic, Intra-country, and Exclusive. We provide the patch-level results on the Open Buildings dataset in Table \ref{tab:open_buildings_results}. 

We see that among the three experiment settings, Intra-country gives the highest Pearson's $r^2$ of 0.971 because the model is tested on in-domain data and the region is small. We also observe that the proposed multi-node quantile regression model generalizes well to regions not seen during training. Speicifcally, under the Exclusive setting, the proposed method achives an $r^2$ as high as 0.962 even with the test regions removed from the training set (see Table \ref{tab:spacenet_results}).

Furthermore, the proposed model generalizes to regions outside of the African continent. We evaluate our method on SpaceNet7 tiles from South America and provide the results in Table \ref{tab:spacenet_results}. We observe that the proposed model achieves comparable or superior performance when evaluated on Brazil, Peru, and Mexico compared with the baselines. This generalization to a different continent indicates that our model could potentially be applied to the globe for collecting building coverage statistics, while using only publicly available low-resolution satellite imagery.

\begin{figure}[t]
\begin{center}
\includegraphics[width=1.0\linewidth]{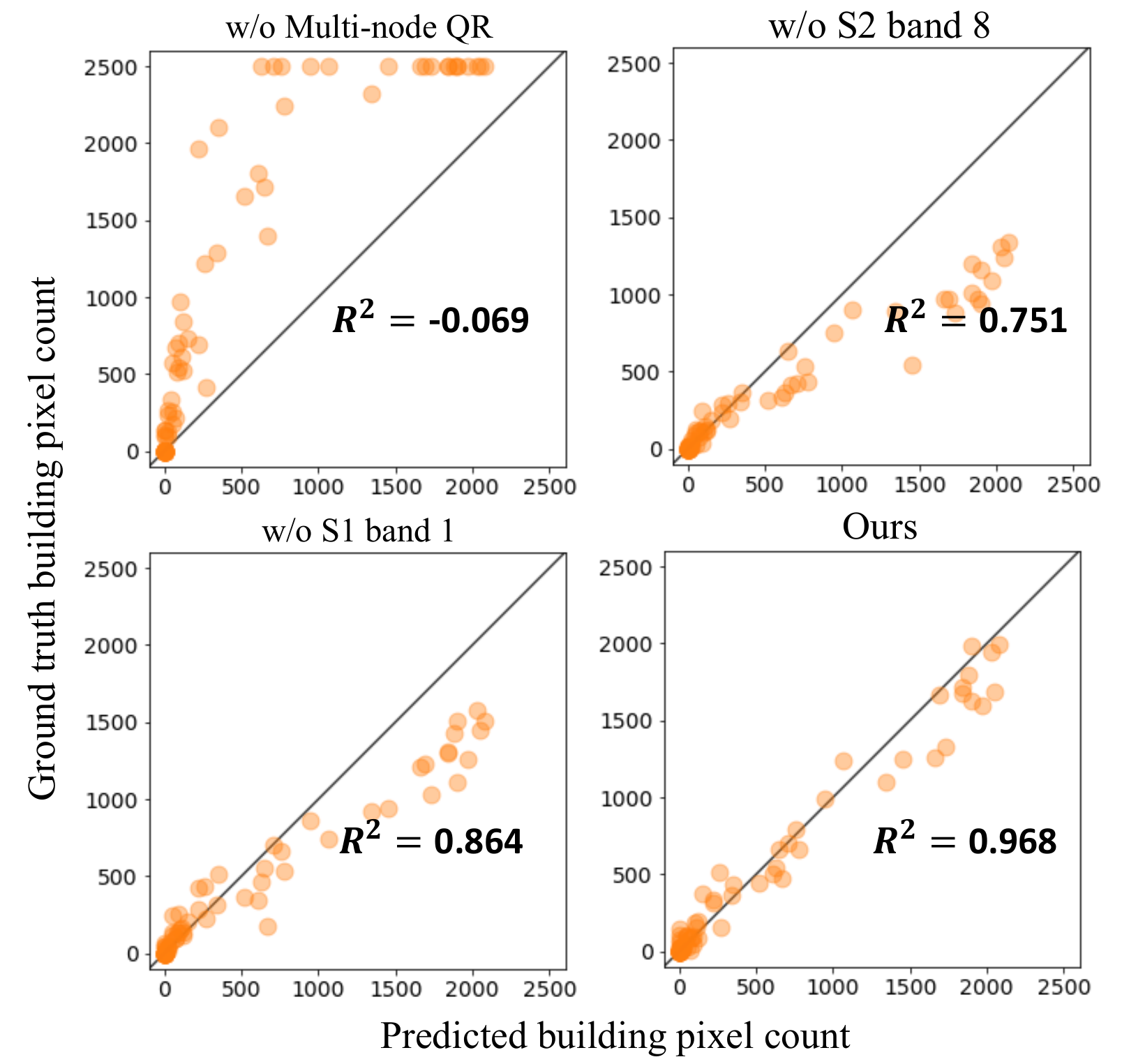}
\end{center}
   \caption{Scatter plots of patch-level predicted number of building pixels against the ground truth from the ablation studies. All models are trained on Africa and tested in the Brazil tile. Removing the multi-node quantile regression or any of the non-RGB bands makes the model be prone to overestimation or underestimation.
   }
\label{fig:pred_vs_gt}
\end{figure}

\subsection{Ablation Studies}
In this part, we carry out ablation studies on 1) the incorporation of S1 band 1 and S2 band 8 as input and 2) the multi-node quantile regression, and provide the results in Table \ref{tab:spacenet_results}.

\subsubsection{Multi-node Quantile Regression}
To see the effect of having multiple output nodes for different quantiles, we compare the performance of the multi-node quantile regression model with the single-node model trained on the L1 objective.
We provide the ablation study results on SpaceNet7 in Table \ref{tab:spacenet_results} (see row ``w/o multi-node QR''). 
We observe that the proposed multi-node model outperforms the single-node model by a large margin on all test regions. 
In addition, as shown in the scatter plots for ground truth VS. predicted building pixel counts (Figure \ref{fig:pred_vs_gt}), the model tends to overestimate the building coverage without the multi-node quantile regression. We speculate that having multiple output nodes improve performance because the pinball losses from the 0.1 and 0.9 quantile nodes help bound the predictions.

\subsubsection{Multi-Spectral Bands}
We conduct ablation studies on the two non-RGB multi-spectral bands --- i.e. S1 band 1 and S2 band 8 --- and provide the ablation study results evaluated on SpaceNet7 in Table \ref{tab:spacenet_results} . We observe that at both patch- and tile-level, incorporating S1 band 1 and S2 band 8 boosts the performance. 
From the scatter plots in Figure \ref{fig:pred_vs_gt}, we see that removing either S1 band 1 or S2 band 8 makes the model underestimates at patch-level. This suggests that the non-RGB bands provide additional information that helps correct the model's tendency to under- or over-estimate.

\section{Discussion and Social Impact}
United Nations' Sustainable Development Goals (SDGs) present an urgent call for action in all countries and collaboration between different sectors of policy-making for a more sustainable development \cite{unsdg}. However, relevant data for informing the decision makers and organizations are often lacking or infrequently collected, especially in fast developing countries. Building coverage is an important socioeconomic indicator and also helps predict other indicators. 
In this paper, we develop a framework for estimating building coverage using only free low-resolution satellite imagery from Sentinel-1 and Sentinel-2. The proposed multi-node quantile regression model yields fairly accurate estimates and generalizes well to unseen countries and continents.

This paper offers a cost-efficient and generalizable way to collect global building coverage statistics, accelerating the progress towards multiple SDGs. For example, building coverage helps predict the level of economic development, which informs policymakers' decisions for alleviating poverty (SDG 1 No Poverty), allocating infrastructure resources (SDG 9 Industry, Innovation and Infrastructure), and building more sustainable cities (SDG 11 Sustainable Cities and Communities). In addition, building coverage provides important information about the interaction between human and environment, including the monitoring of agriculture to reduce hunger (SDG 2 Zero Hunger) and climate measurement (SDG 13 Climate Action). 

Furthermore, the proposed method could potentially be applied to track changes of building coverage for different regions in the world, assisting existing efforts for this. For example, United Nations' World Urbanization Prospects report provides estimates and projections of urban and rural data, including population and area, throughout the time \cite{urbanization}. The proposed method could be applied to derive building coverage statistics as soon as satellite images are updated (the low-resolution satellite imagery are updated on a weekly basis). As building coverage is highly correlated with or can be used to derive other values like building density, urban area, and population, our method could potentially aid the efforts to track urban development.

This paper demonstrates the viability of using low-resolution free satellite imagery for estimating building coverage statistics over a large geography. Future research could explore how the model can be applied to classify urban and rural areas, and estimate population and poverty levels.

\bibliography{aaai23}
\clearpage
\appendix
\section{Appendix A: Datasets}

\paragraph{Sentinel-1} provides free global Synthetic Aperture Radar (SAR) imagery available with 9 bands at ground sampling distance (GSD) of 10m. The revisit time of the Sentinel-1 satellite is 12 days, meaning that the images are updated frequently. 
We include band 1, the VV overall mean, as one of the input channels. 
Each Sentinel-1 tile that we download is 2km-by-2km and has the shape $200 \times 200$ pixels. Each tile is then cropped into 16 smaller patches of shape $50 \times 50$ pixels when fed into our model.

\paragraph{Sentinel-2} provides multi-spectral images in from the visible to the shortwave infrared spectral range (SWIR) with the revisit time of 10 days.
The Sentinel-2 satellite imagery contains 13 multi-spectral bands with GSDs ranging from 10 to 60m. 
Besides the RGB channels, we also included the near-infrared (NIR) channel (i.e. band 8) as input. 
Like Sentinel-1, all the Sentinel-2 imagery are downloaded as 2km-by-2km tiles and cropped into 16 patches of size $50 \times 50$-pixel.

\paragraph{Open Buildings}
contains 516M building footprints across 43 African countries that cover 64\% of the continent. The building footprints were detected using state-of-the-art segmentation models. The Open Buildings data is originally in the format of polygons labeled with three confidence intervals of buildings presence - 0.6 to 0.65, 0.65 to 0.7, and greater than 0.7. For our purpose, we download the Open Buildings data as high-resolution rasters with 0.5m GSD and two bands. 
Band 1 demonstrates the model confidence that a building is located in the region with the confidence interval preserved from the original polygon data. A band 1 value of zero indicates no building presence. Band 2 is a reclassification of the confidence scores into four buckets. 

We use band 1 to derive the binary label of \textit{building} and \textit{non-building}. Specifically, we first downsample the rasters to 10m GSD to match the resolution of the Sentinel-1 and Sentinel-2 input images. Then, we convert the continous-valued band 1 into a binary mask by treating all pixels with a non-zero value as a building pixel -- i.e. a pixel that contains buildings. Like Sentinel-1 and Sentinel-2, the downsampled binary mask is cropped into $50 \times 50$-pixel patches. For each smaller patch, we use the number of building pixels as the labels for training and testing.

\subsection{Appendix B: Baselines}
\begin{figure}[t]
\begin{center}
\includegraphics[width=1.0\linewidth]{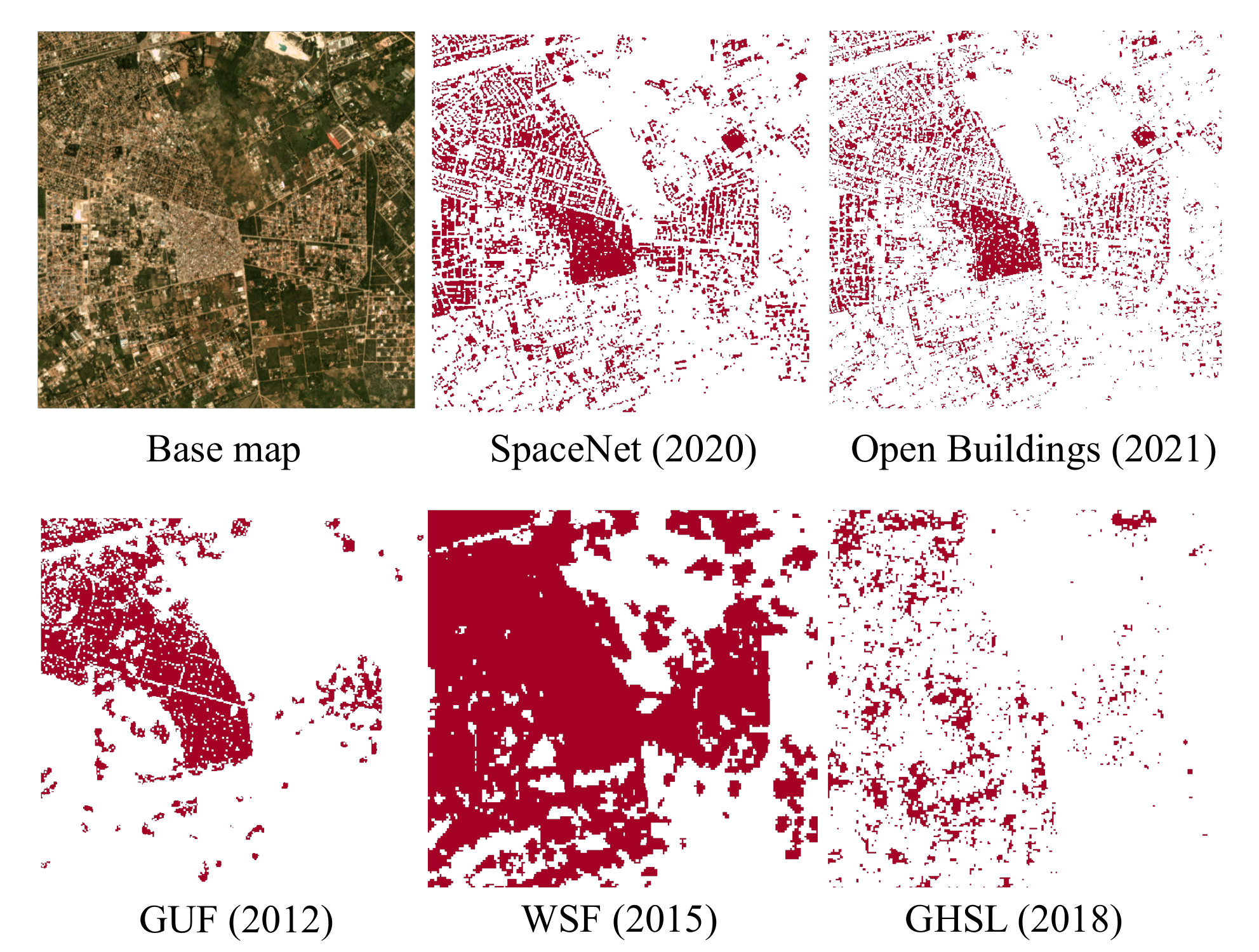}
\end{center}
   \caption{Binary settlement maps in Zambia (1 = red building pixel; 0 = white non-building pixel) from different sources. The years of collection for the settlement maps are provided in the parentheses.}
\label{fig:benchmark_vis}
\end{figure}

Figure \ref{fig:benchmark_vis} shows the visualization of different benchmarks described below and datasets we used in experiments.

\paragraph{Gloal Urban Footprint (GUF)} is a worldwide mapping of human settlement patterns in the form of binary masks available in 12m GSD. The building footprints are derived from satellite images of TerraSAR-X and TanDEM-X from 2011 to 2012.

\paragraph{Global Human Settlement Layer (GHSL)} was collected from backscattered information of Sentinel-1 images. The building map is available as binary masks at 20m GSD.

\paragraph{World Settlement Footprint (WSF)}
is a binary mask of global human settlements available in 10m GSD. The map was derived from Sentinel-1 and Landsat-8 satellite imagery in year 2015.

\begin{figure}[t]
\begin{center}
\includegraphics[width=1.0\linewidth]{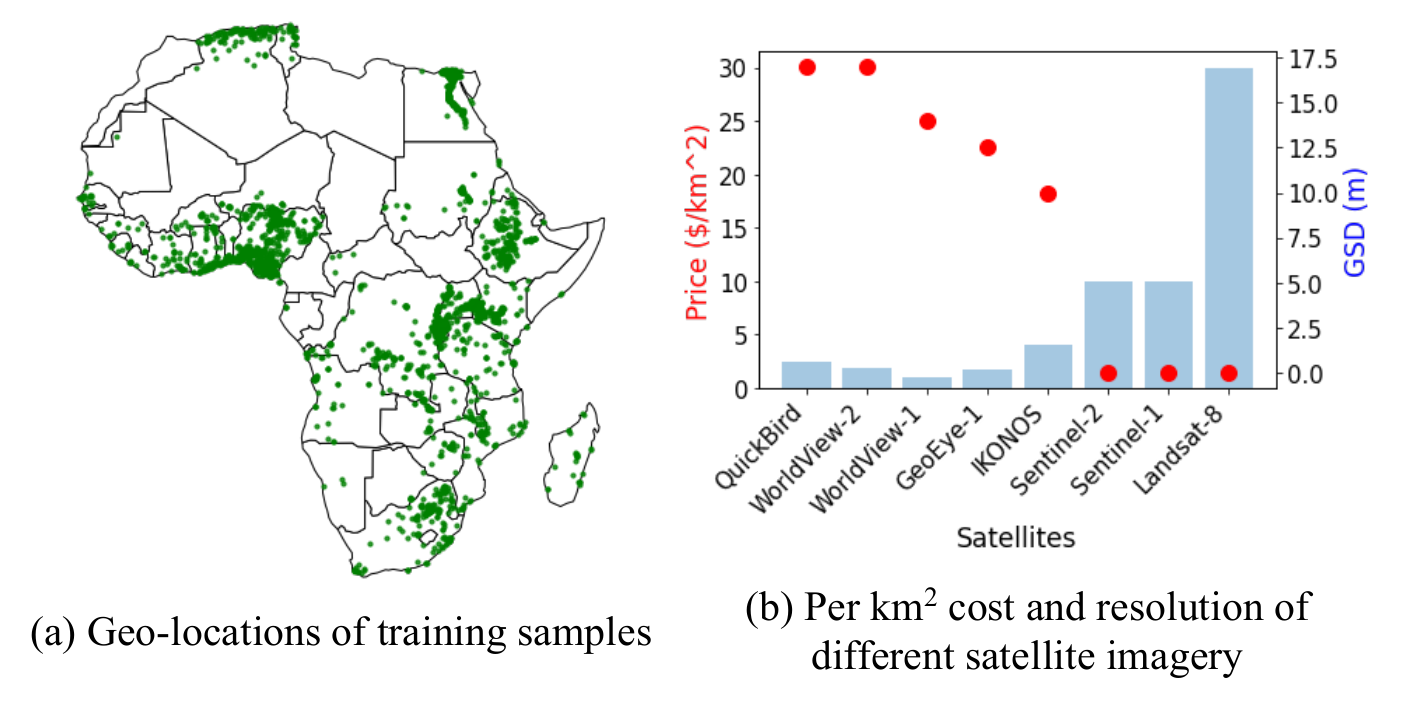}
\end{center}
   \caption{(a) The geo-locations of our training samples, all of which are from Africa. (b) High-resolution satellite imagery are expensive, while many lower-resolution ones are publicly available.}
\label{fig:train}
\end{figure}

\section{Appendix C: Experiments}
\subsection{Model Implementation}
We modify the ResNet18 architecture to incorporate multi-node quantile regression. Specifically, we replace the final fully-connected (FC) layer in ResNet18 with two FC layers, each followed by a ReLU activation. We set the number of output channels (i.e. nodes) to be $K$, which is the number of quantiles the model predicts. For all models in this paper, we use $K=3$ and predict the quantiles $\{0.1, 0.5, 0.9\}$. Each channel corresponds to a quantile and the corresponding pinball loss is computed using the output of that particular channel. To prevent the model from overfitting, we add dropout layers after each ResNet block, which we found empirically that it helps the model generalize to unseen regions.

\subsection{Training Details}
The training samples' geo-locations are shown in Figure \ref{fig:train} (a). 
For all models in the experiment section, we train them on 15000 samples for 1200 epochs with a learning rate of 0.002. The models have fully converged at the point where we end training. 

\subsection{Temporal Experiments}

\subsubsection{Evaluation}
To evaluate our model's ability to track temporal changes in building coverage, we run the model on satellite images from 2016 and 2021.\footnote{Sentinel-2 mission started in 2014.} Specifically, we chose four cities with various levels of development from four different continents -- Dhaka, Kampala, Athens, and Boston -- and downloaded all images covering the cities. Then, we compute the building coverage growth rate over the 5-year interval for the chosen cities using the model trained on 15,000 African images for 1200 epochs. 

One challenge of temporal evaluation is the lack of building coverage ground truth data. To get a sense of how well the proposed method tracks temporal changes in building coverage, we use population change as a proxy for building coverage change.
Specifically, we assume that population and building coverage grows together, as more people requires more buildings. However, we only use population growth rate as a rough reference for the relative level of developments of different cities. 

\subsubsection{Results}
Tracking changes in building coverage across time allows us to understand the urban development of a region, especially in cities or urban areas. We show that the proposed model can be used to track building coverage changes in regions of different levels of development and provide the temporal experiment results in Figure \ref{tab:temporal_results}. 

\begin{table}[t]
\centering
\begin{tabular}{c|c|c}
\Xhline{3\arrayrulewidth}
\textbf{Region}    & \textbf{Population (\%)}   & \textbf{Building (\%)}\\ \hline
Dhaka  & 19.23 &  29.68    \\ 
Kampala   & 28.18   &  7.28    \\
Athens  & -0.19 & 1.55   \\
Boston & 4.45 &  -9.42   \\ \Xhline{3\arrayrulewidth}
\end{tabular}
\caption{Temporal change experiment results on four cities of different levels of development. The second and third are percentage change from 2016 to 2021. All results are collected from the model trained with 15,000 samples in Africa for 1200 epochs.}
\label{tab:temporal_results}
\end{table}

\end{document}